\documentclass{article}

\usepackage{ijcai13}
\usepackage{times}

\usepackage{amsmath}
\usepackage{amssymb}
\usepackage{amsthm}
\usepackage{xspace}
\usepackage{enumerate}
\usepackage{booktabs}
\usepackage{url}

\usepackage{graphicx} 

\newcommand{\tuple}[1]{\langle#1\rangle}
\newcommand{\Omit}[1]{}
\newcommand{\tup}[1]{\langle #1 \rangle}
\newcommand{\pair}[1]{\langle #1 \rangle}

\newcommand{\CHECK}[1]{\textcolor{red}{\bf CHECK: #1}}

\newcommand{\sci}[2]{#1\textsc{e}#2}

\newtheorem{theorem}{Theorem}

\newtheorem{definition}[theorem]{Definition}
\newtheorem{definition-and-theorem}[theorem]{Definition and Theorem}

\usepackage{relsize}
\newcommand{\join}{\mathlarger{\Join}}
\DeclareMathOperator{\bigjoin}{\mathlarger{\mathlarger{\mathlarger{\Join}}}}

\newcommand{\citeay}[1]{\citeauthor{#1}~[\citeyear{#1}]}

\begin{document}

\title{Causal Belief Decomposition for Planning with Sensing: \\
       Completeness Results and Practical Approximation}

\author{Blai Bonet \\
        Universidad Sim\'on Bol\'{\i}var \\
        Caracas, Venezuela \\
        {\normalsize\url{bonet@ldc.usb.ve}}
\And
        Hector Geffner \\
        ICREA \&  Universitat Pompeu Fabra \\
        08018  Barcelona, SPAIN \\
        {\normalsize\url{hector.geffner@upf.edu}}
}
\maketitle

\begin{abstract}
Belief tracking is a basic problem in planning with sensing.
While the problem is intractable, it has been recently shown
that for both \emph{deterministic} and \emph{non-deterministic}
systems expressed in compact form, it can be done in time and
space that are exponential in the problem \emph{width}.
The width measures the maximum number of state variables that
are all \emph{relevant} to a given precondition or goal. 
In this work, we extend this result both theoretically and practically. 
First, we introduce an alternative decomposition scheme and
algorithm with the same time complexity but  different
completeness guarantees, whose space complexity is much smaller: 
exponential in the \emph{causal width} of the problem
that measures the number of state variables that are 
\emph{causally relevant} to a given precondition, goal, or
observable. Second, we introduce a fast, meaningful, and powerful 
approximation that trades completeness by speed, and is both
\emph{time and space exponential in the problem causal width}.
It is then shown empirically that the algorithm combined with
simple heuristics yields state-of-the-art real-time performance
in domains with high widths but low causal widths such as
Minesweeper, Battleship, and Wumpus.
\end{abstract}

\section{Introduction}
Planning with sensing is a search problem in belief space that
involves two tasks: keeping track of beliefs and selecting the
action to do next.
In this paper, we consider the first task in  a logical, non-probabilistic setting, where \emph{beliefs} stand
for \emph{sets of states}. While belief tracking for problems
expressed in compact form is computationally intractable,
it has been recently shown that for \emph{deterministic} problems, 
the problem can be solved in time and space that are exponential in
a problem \emph{width} parameter that in many of the existing
benchmarks is bounded and small \cite{palacios:jair09,albore:ijcai09}.
The same result has been extended to \emph{non-deterministic}
problems where the task of keeping track of global beliefs $b$ over
all problem variables is replaced by the task of keeping track of
local beliefs $b_X$ for each precondition and goal variable $X$,
each of which involves the variables that are \emph{relevant} to $X$.
The width $w(X)$ of a variable $X$ is the number of
variables that are \emph{relevant} to $X$, and the
\emph{width of the problem}, $w(P)$, is  the maximum number of
variables that are all relevant to a precondition or goal variable.
The algorithm that tracks beliefs in time and  space that are exponential
in the problem width is called \emph{factored belief tracking}
\cite{bonet:bel-aaai2012}.

A limitation of these accounts is that many meaningful domains
have large, unbounded widths.
\Omit{
This not surprising as  belief tracking in many such  domains is indeed intractable 
yet we want to be able to track beliefs efficiently in such domains even if not 100\% accurately. 
Belief tracking scheme over a compact representation is 
said to be \emph{sound} when the literals $X=x$ and $X \not= x$ that are found
to be true after an execution, must indeed be true, and it is \emph{complete}
when the literals  $X=x$ and $X \not= x$ that must be true after an execution,
are found to be true. From a practical point of view, we are thus looking for
belief tracking algorithms that are sound, fast, and powerful, but not necessarily 
complete. Moreover, we want to understand the source of  incompleteness, 
and obtain useful information about the relative number of states possible
after an execution, where a literal $X=x$ holds, as opposed to literals 
$X=x'$, where $x$ and $x'$ are two possible values for variable $X$. 
These ratios encode relative probabilities under uniformity assumptions, 
and they are needed in problems like Minesweeper or Battleship,
where often  the choices are  not sure bets. 
}%
In this work, we aim to extend these results so that such problems can
be handled effectively. For this, we introduce an alternative decomposition
scheme and belief tracking algorithm with the same time complexity as
factored belief tracking but whose space complexity is often much smaller:
exponential in the \emph{causal width} of the problem.
The causal width measures the number of state variables that are all
\emph{causally relevant} to a given precondition, goal, or observable.
We also  determine the conditions under which the new belief tracking
scheme is complete, and use the new decomposition to define an
incomplete but meaningful and powerful approximation algorithm that is
practical enough, as it is both \emph{time and space exponential
in the problem causal width.} 

The paper is organised as follows. After reviewing the basic model and   notions from 
\citeay{bonet:bel-aaai2012},  we  introduce the  new notion of causal width, 
a new decomposition  scheme,  and the new algorithm. We then  test  the  algorithm
experimentally.

\section{Planning with Sensing} 

We review the  standard model for planning with sensing,
and a language for representing these models in compact form. 

\subsection{Model}

The model for \emph{planning with sensing} is a simple extension of
the model for conformant planning where a goal is to be achieved
with certainty in spite of uncertainty in the initial situation or
action effects. The model for conformant planning is characterized by a tuple
$\mathcal{S}= \tuple{S,S_0,S_G,A,F}$ where 
$S$ is a finite state space, $S_0$ is a non-empty set of possible initial states, 
$S_G$ is a non-empty set of goal states, $A$ is a set of actions with $A(s)$ denoting the actions
applicable in state $s \in S$, and $F$ is a non-deterministic transition function such that $F(a,s)$
 denotes the non-empty set of possible successor states that follow action   $a$ in state $s$, $a \in A(s)$.

\Omit{
\begin{enumerate}[\ $\bullet$]
\item $S$ is a finite state space,
\item $S_0$ is a non-empty set of possible initial states, $S_0 \subseteq S$, 
\item $S_G$ is a non-empty set of goal states, $S_G \subseteq S$,
\item $A$ is a set of actions, with $A(s)$ denoting the sets of actions
  applicable in $s \in S$, and 
\item $F$ is a non-deterministic transition function such that $F(a,s)$
  denotes the non-empty set of possible successor states that follow action
  $a$ in state $s$, $a \in A(s)$.
\end{enumerate}
}
\Omit{
\noindent A solution to a conformant model is an action sequence that
maps each possible initial state into a goal state. More precisely,
$\pi= a_0, \ldots, a_{n-1}$ is a conformant plan if for each possible
sequence of states $s_0, s_1, \ldots, s_{n}$ such that $s_0 \in S_0$ 
and $s_{i+1} \in F(a_i,s_i)$, $i=0,\ldots,n-1$, then the action $a_i$
is applicable in $s_i$ and $s_{n}$ is a goal state. 
}

Conformant planning can be cast as a path finding problem over \emph{beliefs}
defined as the sets of states that are deemed possible at any one point 
\cite{bonet:aips2000}. The initial belief $b_0$ is $S_0$, and the belief $b_a$
that results from an action $a$ in a belief state $b$ is:
\begin{equation}
\label{eq:ba}
b_a = \{s' \, | \, \text{there is $s \in b$ such that $s' \in F(a,s)$} \}\,, 
\end{equation}
where it is assumed that the action $a$ is applicable at each state $s$ in $b$.
%
\emph{Contingent planning} or \emph{planning with sensing} is planning with
both uncertainty and feedback. The model for contingent planning is the model
for conformant planning extended with a \emph{sensor model}. A sensor model
is a function $O(s,a)$ that maps state-action pairs into observations tokens $o$.
The expression $o \in O(s,a)$ means that token $o$ is a possible observation
when $s$ is the true state of the system and $a$ is the last action done.
%
Executions in the contingent setting are sequences of
action-observation pairs $a_0,o_0,a_1,o_1,\ldots$. If $b=b_i$ is the belief
state when the action $a_i$ is applied and $o_i$ is the token that is observed,
then the belief $b_a$ after the action $a=a_i$ is given by Eq.\,\ref{eq:ba},
and the belief $b_{i+1} = b_a^o$ that follows from observing the token $o$ is:
\begin{equation}
\label{eq:bao}
b^o_a = \{s \,|\, \text{$s\in b_a$ and $o\in O(s,a)$} \} \,.
\end{equation}
An execution $a_0,o_0,a_1,o_1,\ldots$ is deemed \emph{possible} in $P$
if starting from the initial belief $b_0$, each action $a_i$ is applicable
at the belief $b_i$ (i.e., $a_i\in A(s)$ for all $s\in b_i$), and
$b_{i+1}$ is non-empty. The contingent model is similar to 
POMDPs but with uncertainty encoded through sets rather 
than probability distributions.


\Omit{
In both cases, the action selection strategy can be expressed 
as a \emph{partial} function $\pi$ over beliefs, called a \emph{policy},
such that $\pi(b)$ is the action to do in belief $b$. 
The function is partial because it has to be defined only over the initial belief
$b_0$ and the non-goal beliefs $b$ that can be reached with $\pi$ from $b_0$.
Such partial partial policies can be represented as graphs where
the nodes are the reachable beliefs.
}

\subsection{Syntax}

Syntactically, conformant problems can be expressed in \emph{compact form}
through a set of \emph{state variables}, which for convenience we assume
to be \emph{multi-valued}. More precisely, a conformant planning problem is a tuple $P = \tup{V,I,A,G}$
where $V$ stands for the problem variables $X$, each one with a finite 
domain $D_X$, $I$ is a set of clauses over the $V$-literals defining
the initial situation, $A$ is a set of actions, and $G$ is a set of $V$-literals
defining the goal. Every action $a$ has a precondition $Pre(a)$, given by a set of $V$-literals, 
and a set of conditional effects $C \rightarrow E_1 | \ldots | E_n$, where $C$
and each $E_i$ is a set (conjunction) of $V$-literals.
The conditional effect is \emph{non-deterministic} if $n>1$;
else it is deterministic. A conformant problem $P = \tup{V,I,A,G}$ defines a conformant model
${\cal S}(P) = \tuple{S,S_0,S_G,A,F}$, where $S$ is the set of 
valuations over the variables in $V$, $S_0$ and $S_G$ are the set of
valuations that satisfy $I$ and $G$, $A(s)$ is the set of
operators whose preconditions are true in $s$, and $F(a,s)$ is 
determined by the conditional effects of $a$ in the standard way.

Contingent problems can be described by extending the syntactic
description of conformant problems with a compact encoding of the
\emph{sensor model}.
For this, we assume a set $V'$ of observable multi-valued variables $Y$,
not necessarily disjoint with the set of state variables $V$ (i.e., some state
variables may be observable), and formulas $W_a(Y=y)$ over a subset of
state variables, for each action $a$ and each possible value $y$ of each
observable variable $Y$. The formula $W_a(Y=y)$ encodes the states over
which the observation $Y=y$ is possible when $a$ is the last action. 
\Omit{
The formulas $W_a(Y=y)$ for different $y$'s in $D_Y$ must be
logically \emph{exhaustive}, as every state-action pair must yield
some observation $Y=y$. If in addition, the formulas $W_a(Y=y)$ for
different $y$'s are logically \emph{exclusive}, every state-action
pair gives rise to a single observation $Y=y$, and the sensing over
$Y$ is deterministic. 
If a state variable $X$ is observable, then $W_a(X=x)$ is just the formula `$X=x$'.
}%
A contingent problem is  a tuple $P=\tup{V,I,A,G,V',W}$ that defines a
contingent model that is given by the conformant model $\tuple{S,S_0,S_G,A,F}$ 
determined by the first four components in $P$, and the sensor model
$O(s,a)$ determined by the last two components, where $o \in O(s,a)$
iff $o$ is a valuation over the observable variables $Y\in V'$ such that
$Y=y$ is true in $o$ only if the formula $W_a(Y=y)$ in $W$ is true in
$s$.

\Omit{
As an illustration, if $X$ encodes the position of an agent, and $Y$ encodes
the position of an object that can be detected by the agent when $X=Y$, we
can have an observable variable $Z \in \{Yes,No\}$ with model $W_a(Z=Yes)$
given by $\vee_{l \in D} (X=l \land Y=l)$, and $W_a(Z=No)$ given by the negation
of this formula. Here $D$ is the set of possible locations and $a$ is any action.
This will be a deterministic sensor. A non-deterministic sensor could be used if,
for example, the agent cannot detect the presence of the object at certain
locations $l \in D'$. For this, it suffices to set $W_a(Z=No)$ to the disjunction
of its previous formula and $\vee_{l \in D'} (X=l \land Y=l)$. 
}

For simplicity and without loss of generality, we make three simplifying  assumptions
\cite{bonet:bel-aaai2012}; namely, that the initial situation  $I$ is given
by a set of literals, that non-deterministic conditional effects
involve just one variable in their heads, and that every observable variable
is relevant to a variable appearing in some precondition or goal. 
If these assumptions are not true for a problem, they can be enforced by means of simple, 
equivalence-preserving  transformations.\footnote{A longer version of the paper provides
  the proofs and necessary details.}
Extensions of this basic  language featuring \emph{defined variables} and
\emph{state constraints} are discussed below.

\Omit{
\footnote{***1. Add somewhere, perhaps above: Without loss of generality we assume
that the initial situation $I$ contains just literals. 
*** 2. All observations must be relevant to some precondition or goal;
else can be eliminated. This is why no obs $X$ needed.
Include this in assumptions. ***}
\footnote{
***** To check: how to deal with non-det effects $true \rightarrow x \land y | \neg x \land \neg y$,
and $z \rightarrow x \land y | \neg x \land \neg y$. In AAAI-2012, the three vars will be 
in same beam if $y$ causally relevant to an obs, as in such a case, $y$ is evidentially relevant
to $z$. But potential problem: looks that $z$ is also evidentially relevant to $y$, as $z$ and $y$ 
causally relevant to obs. Perhaps no need to be formal about evidentially relevant, just 
causal relevance and plain relevance. Adopt JAIR: evidentially relevant not causally closed 
and requires causal relevance in other directions; see JAIR draft.  Also about factorization of $I$.
***}
\footnote{\CHECK{We have more assumption in JAIR draft, some of which (besides unit clauses in $I$)
are necessary and not mentioned in the paper}}
}


\section{Belief Tracking for Planning}

A real execution is an interleaved sequence of actions and observations
$a_0,o_0,a_1,o_1,\ldots$, where $a_i$ is an action from the problem
and $o_i$ is a full valuation over the observable variables. 
We will find it convenient, however, to consider \emph{generalized
executions} $a_0,o_0,a_1,o_1,\ldots$ where the $o_i$'s
denote partial valuations, and in particular, \emph{observation
literals} $Y=y$. Any real execution can be expressed as a
generalized execution provided that the observation literals
that arise from the same (full) observation are separated by dummy NO-OP actions
with no effects. The \emph{belief tracking for planning} problem can then be
expressed as follows:

\begin{definition}
The \emph{belief tracking for planning (BTP) problem} is the
problem of determining whether a generalized execution $\tau: a_0,o_0,a_1,o_1,\ldots,a_k,o_k$ is possible
for a problem $P$, and whether it achieves the goal.
\end{definition}

In other words, for a planner to be complete, it just needs to determine 
which observations are possible after an execution, which actions are applicable, and whether the 
goal has been achieved. There is no need, on the other hand, to determine whether an arbitrary formula 
is true after an execution. This is an important distinction which is not exploited
by the baseline algorithm for BTP, which is called \emph{flat belief tracking} 
and is the result of the iterative application of Equations~\ref{eq:ba} and \ref{eq:bao}.
The complexity of flat belief tracking is exponential
in the number of state variables $|V|$. Often, however, the value
of some variables is not uncertain and such variables do  not add up to the
complexity of tracking beliefs.  We call such variables \emph{determined}.
Formally, a \emph{set} of state variables $X$ is \emph{determined} when
they are all initially known, they appear in the heads of deterministic
effects only, and the variables appearing in the body of such effects,
if any, are in the set as well. The set of determined variables for
problem $P$ is the maximal set of variables that is determined.

The BTP problem, however, remains intractable in the worst case:

\begin{theorem}
\label{thm:complexity}
The belief tracking for planning (BTP) problem is NP-hard and coNP-hard.
\end{theorem}

\section{Width and Factored Belief Tracking}

Factored belief tracking \cite{bonet:bel-aaai2012} improves flat belief tracking 
by focusing on the beliefs that are strictly necessary for solving BTP,
and by exploiting the structure of the problem.
For this, a variable $X$ is regarded as an \emph{immediate cause} of
a variable $X'$ in a problem $P$, written $X \in Ca(X')$, if
$X \neq X'$, and either $X$ occurs in the body $C$ of a
conditional effect $C \rightarrow E_1|\cdots|E_n$ such that
$X'$ occurs in a head $E_i$, $1\leq i\leq n$, or $X$ occurs
in a formula $W_a(X'=x')$ for an observable variable $X'$
and value $x'\in D_{X'}$. \emph{Causal relevance} is  the transitive closure of
this relation:

\begin{definition}
$X$ is \emph{causally relevant} to $X'$ in $P$ if $X=X'$, $X \in Ca(X')$, or 
$X$ is causally relevant to a variable $Z$ that is causally relevant to $X'$.
\end{definition}

In the presence of observations, however, relevance does not
flow only \emph{causally}, but like in Bayesian Networks
\cite{pearl:book}, it also flows \emph{evidentially:}

\begin{definition}
$X$ is \emph{evidentially relevant} to $X'$ in $P$ if $X'$ is causally
relevant to $X$ and $X$ is an observable variable.
\end{definition}

\noindent Relevance is the transitive closure of the causal and
evidential relations taken together:

\begin{definition}
$X$ is \emph{relevant} to $X'$ if $X$ is causally or evidentially relevant
to $X'$, or $X$ is relevant to a variable $Z$ that is relevant to $X'$.
\end{definition}

\Omit{
For example, $X$ is relevant to $X'$ if $X$ is causally relevant to $Z$
and $Z$ is evidentially relevant to $X'$. From a Bayesian Network perspective,
the notion of relevance encodes `potential dependency' given what may
be observed, using the information that certain variables will not be observed.
The immediate cause relations along with the distinction between state
and observable variables determine a causal graph from which the 
relevance relations can be read. Action preconditions do not add
further links to such a causal graph as action preconditions must be
known with certainty and hence do do not propagate uncertainty,
which is what belief tracking is about. 
}


\noindent The \emph{width of a variable} $X$ and \emph{the width of the problem} are defined 
in terms of the relevance relation:

\begin{definition}
The \emph{width} of a variable $X$, $w(X)$, is 
the number of state variables that are relevant to $X$
and are not determined. The width  of the problem $P$, $w(P)$, is
$\max_X w(X)$, where $X$ ranges over the variables that appear
in a goal or action precondition.
\end{definition}

The result obtained by \citeay{bonet:bel-aaai2012} is an
algorithm that solves the BTP problem in  time and space that are
exponential in the problem width. For this, \emph{factored belief tracking} does not track the global
belief $b$ over all the problem variables after an execution, 
but tracks the `local beliefs' $b_X$ over the state variables that are relevant to each variable $X$ 
appearing in action preconditions or goals. This is sufficient for solving BTP. 
\Omit{
In fact, from such local beliefs, it is possible to determine 
if the precondition or goal literal $X=x$ is true or false, and also
whether an observation literal $Y=y$ is true or false for a variable $Y$
relevant to $X$.\footnote{It is assumed that every state and observable variable in the problem is relevant
to at least some variable appearing in an action precondition or goal. 
If this is not the case, it can be shown that such variables can be
projected away from the problem. \CHECK{This can be added to revised assumption above}.
}}
The \emph{local beliefs} $b_X$ are defined in turn as the \emph{global beliefs}
that result from an execution over  a problem that is like $P$ but with all state variable that are not
relevant to $X$, projected away.
These \emph{projected problems} are defined as:

\begin{definition}
\label{projection}
The projection of problem $P = \langle V,I,A,G,$ $V',W\rangle$ on a subset of state
variables $S \subseteq V$ is $P_S = \langle V_S,$ $I_S,A_S,G_S,$ $V'_S,W_S\rangle$
where $V_S$ is $S$, $I_S$ and $G_S$ are the initial and goal formulas
$I$ and $G$ (logically) projected over the variables in $S$, $A_S$ is $A$
but with preconditions and conditional effects projected over $S$,
$V'_S$ is $V'$, and $W_S$ is the set of formulas $W_a(Y=y)$ in $W$
projected on the variables in $S$.\footnote{The logical projection of formula $F$ over a subset $S$ of
its variables refers to the formula $F'$ defined over the variables in
$S$, such that the valuations that satisfy $F'$ are exactly those
that can be extended into valuations that satisfy $F$ \cite{darwiche:map}.
}
\end{definition}

For convenience, the projection $P_S$ of $P$ where $S$ is the set of state variables relevant to $X$
is abbreviated as $P_X$.  The projections $P_X$ for variables $X$ appearing in action preconditions
or goals ensure three key properties: first, that the real and generalized executions $a_0,o_0,a_1,o_1,\ldots$
that are possible in $P$ are possible in $P_X$; second,
that belief tracking over the projections $P_X$ is time and space exponential 
in $w(P)$, and finally, that the beliefs $b_X$ and $b$ that result
from such executions over $P_X$ and $P$ are \emph{equivalent} over the
set of variables relevant to $X$.
\Omit{ 
\emph{Factored belief tracking} is the algorithm that keeps track of
the local beliefs $b_X$ by performing flat belief tracking over each
of the projections $P_X$:

\begin{definition}
\emph{Factored belief tracking} is the algorithm that tracks the
beliefs $b_X$ that result from a given execution in $P$ by applying
\emph{flat belief tracking} over each  projected problem
$P_X$, for $X$ being a variable appearing in action preconditions
or goals of $P$.
\end{definition}

\noindent The main result in \cite{bonet:bel-aaai2012} is that:
}
\emph{Factored belief tracking} is the algorithm that solves 
the belief tracking problem over $P$ by  keeping track of
the local beliefs $b_X$ for each precondition and goal variable $X$
using \emph{flat belief tracking} over each of the projections $P_X$:

\begin{theorem}[\citeay{bonet:bel-aaai2012}]
\label{thm:comp1}
Factored belief tracking solves the belief tracking problem
for planning in time and space that are exponential in the problem width.
\end{theorem}


\section{Causal vs.\ Factored Decompositions}

While there are many domains that have a bounded and low width
\cite{palacios:jair09,albore:ijcai09}, there  are also many meaningful domains that do not.  The contribution
of this paper is a reformulation of the above results that lead
to a fast, meaningful, and powerful approximation belief tracking
algorithm that applies  effectively to a much larger class of problems. 
For this, we introduce the idea of \emph{decompositions}, cast
factored belief tracking in terms of one such decomposition, and
introduce an alternative decomposition leading to different
algorithms. 

\begin{definition}
A \emph{decomposition} of a problem $P$ is a pair $D=\{T,B\}$, 
where $T$ is a set of variables $X$ appearing in $P$, called the
\emph{target variables} of the decomposition, and $B$ is the
collection of \emph{beams} $B(X)$ associated with such target
variables such that $B(X)$ is a set of \emph{state variables}
from $P$.
\end{definition}

A decomposition $D=\{T,B\}$ maps $P$ into a set of \emph{subproblems} $P^D_X$,
one for each variable $X$ in $T$, that correspond to the \emph{projections}
of $P$ over the state variables in the beam $B(X)$.
The decomposition that underlies factored belief tracking is:

\begin{definition}
The \emph{factored decomposition} $F=\{T_F,B_F\}$ of $P$ is such
that $T_F$ stands for the set of variables $X$ appearing in action
preconditions or goals, and $B(X)$ is the set of state variables
\emph{relevant} to $X$.
\end{definition}

Factored belief tracking is \emph{flat belief tracking} applied to the subproblems 
determined by the \emph{factored decomposition}. The complexity of the algorithm
is exponential in the problem width, which bounds the size of the
beams in the decomposition.  The algorithms that we introduce next are  based
on a different decomposition:

\begin{definition}
The \emph{causal decomposition} $C=\{T_C,B_C\}$ of $P$ is such that
$T_C$ stands for the action precondition, goal, and \emph{observable}
variables in $P$, and $B_C(X)$ is the set of state variables that are
\emph{causally relevant} to $X$.
\end{definition}

The \emph{causal decomposition} determines a larger number of subproblems, as
subproblems are generated also for the observable variables $X$, but
the beams $B_C(X)$ associated with these  subproblems are  smaller, as they
contain only the state variables that are \emph{causally relevant} to
$X$ as opposed to the \emph{relevant} variables. The \emph{causal width} of a problem is defined in terms of the largest
beam in the causal decomposition:

\begin{definition}
The \emph{causal width} of a variable $X$ in a problem $P$, $w_c(X)$,
is the number of state variables that are \emph{causally} relevant to
$X$ and are not determined.  The \emph{causal width} of $P$ is
$\max_{X} w_c(X)$, where $X$ ranges over the target variables in 
the causal decomposition of $P$. 
\end{definition}


The first and simplest belief tracking algorithm defined over
the new  \emph{causal decomposition} is what we call
\emph{Decoupled Causal Belief Tracking} or Decoupled CBT, 
which runs in time and space that are exponential in the 
causal width:

\begin{definition}
Decoupled CBT  is \emph{flat belief tracking} applied independently
to each of the problems $P_X^C$ determined by the \emph{causal decomposition} of $P$. 
\end{definition}

\Omit{
\begin{definition}
\decoupled is \emph{flat belief tracking} applied independently to each of the problems $P_X^C$
determined by the \emph{causal decomposition} of $P$.
\end{definition}

\noindent \decoupled runs in time and space that are exponential in the problem \emph{causal width:}

\begin{theorem}
\decoupled runs in time and space that are exponential in the problem causal width.
\end{theorem}
}

Since causal width is never greater than width and is often much
smaller, Decoupled CBT will run  much faster than factored belief tracking
in general. This, however, comes at a price,  that we express using 
relational operators for \emph{projections} and \emph{joins}, exploiting that
beliefs are tables or relations  whose rows are states, and whose 
columns are  variables.\footnote{%
For example, if $b$ contains the valuations (states) $X=1,Y=1$ and $X=2,Y=2$,
the projection $\Pi_{\{X\}} b$ will contain the valuations $X=1$ and $X=2$. Likewise, if $b'$ 
contains $Y=1,Z=1$ and $Y=1,Z=2$,  the join $b \, \join \,  b'$ will  contain $X=1,Y=1,Z=1$ and $X=1,Y=1,Z=2$.}
The subproblem of $P$ determined for a target variable $X$ in the \emph{causal} decomposition is denoted as $P_X^C$; i.e., $P_X^C = P_{B_C(X)}$.

\begin{theorem}
\label{thm:sound-incomplete}
Decoupled CBT is sound and runs in time and space that are exponential
in $w_c(P)$ but it is not complete; i.e., for any target variable $X$
in the causal decomposition, if $b$ and $b_X$ are the beliefs resulting
from an execution on $P$ and $P_X^C$ respectively,  then
$b_X \supseteq \Pi_{B_C(X)} b$ is necessarily true, but
$b_X \subseteq \Pi_{B_C(X)} b$ is not.
\end{theorem}

One  reason for the incompleteness of Decoupled CBT is that the beliefs $b_X$ associated
with different target variables $X$ are not  independent. Indeed, a variable $Z$ may appear in two beams of the causal decomposition,  with some  variables relevant
to $Z$ in  one beam, and other  variables relevant to $Z$ in the other. 
In the factored decomposition this cannot happen, as all the variables
that are relevant to $Z$ will appear in all the beams that contain $Z$.
In the causal decomposition, beams are kept small by not closing them
with the relevance relation, but as a result, the beliefs over such beams
are not independent.

\section{Causal Belief Tracking}


In \emph{causal belief tracking}, the local beliefs are not tracked independently over 
each of the subproblems $P_X^C$  of the causal decomposition; rather, the local beliefs 
are first progressed and filtered \emph{independently}, but then are merged and projected back onto each
of the  beams, making them consistent with each other. This consistency operation is expressed
using the projection and join operators:

\Omit{.  While in \decoupled, the belief
$b_{X}^{i+1}$ at time $i+1$ for the subproblem $P_X^C$ is
obtained from the belief $b_X^{i}$ by setting $b_X^{i+1}=b^o_a$,
where $b_a^o$ is given by Eqs.\,\ref{eq:ba} and \ref{eq:bao}
with $b=b_X^i$, $a=a_i$ and $o=o_i$, in \full, it is
obtained as follows:
}

\begin{definition}
\label{def:consistent}
\emph{Causal belief tracking} is the belief tracking algorithm that operates on the
causal decomposition $C=\pair{T_C,B_C}$, setting the beliefs
$b_X^0$ at time $0$ over each beam $B_C(X)$, $X \in T_C$, to
the projection of the initial belief  over the beam, and the successive beliefs $b_X^{i+1}$ as:
\begin{equation}
\label{eq:consistent}
b_X^{i+1}\! =\! \textstyle\prod_{B_C(X)}\! \bigjoin\{ (b^i_{Y})_a^o :\! \text{$Y$ in $T_C$ and relevant to $X$}\}
\end{equation}
where $a=a_i$ and $o=o_i$ are the action and obs.\ at
time $i$ in the execution, and $(b^i_Y)_a^o$ is $b_a^o$ from Eqs.\,\ref{eq:ba}--\ref{eq:bao}
with $b=b^i_Y$. 
\end{definition}



Progressing and filtering the local beliefs in the causal decomposition is 
time and space exponential in the problem \emph{causal width},
but the consistency operation captured by the join-project operation 
in \eqref{eq:consistent} is time exponential in the number of state variables that
are relevant to $X$. As a result:

\begin{theorem}
CBT  is  \emph{space}  exponential in the problem
\emph{causal width}, and \emph{time}  exponential in the
problem \emph{width}.
\end{theorem}

CBT is sound but still \emph{not} complete. 
However, the range of problems for which CBT  is complete, unlike
Decoupled CBT, is large and meaningful enough, and it includes for
example the three domains to be considered in the experimental
section: Minesweeper, Battleship and Wumpus.
We express the completeness conditions for CBT by introducing the
notions of \emph{memory variables} and \emph{causally decomposable problems.}  
A \emph{state variable} $X$ is a \emph{memory variable} in a problem $P$ when the value $X^k$ of the variable at any time point $k$
can be \emph{determined uniquely} from an observation of the  value of the variable $X^i$ at any other time point $i$, 
the actions done in the execution, and the initial belief state of the problem. 
Thus, \emph{static variables} are memory variables, as much as variables that are \emph{determined} (Section~3). For the first, $X^k=X^i$,
while for the second, $X^k$ is determined by the initial belief and the actions before time $k$.
All the variables in  \emph{permutation domains}  \cite{amir:filtering} are also memory variables.
\Omit{
\CHECK{Ok. I kind of understand what your are trying to say and it 
is correct. However, it is not fully clear because the variable $X$
does not need to be known (i.e., has a unique valuation in the beam),
yet the wording seems to imply that $X$ has a definite values at times
$i$ and $k$. What you mean (I guess because is correct in such a case)
is that the value of $X$ at time $k$ along any \emph{state trajectory}
is determined by the value of $X$ (and its causal ancestors)  at time
$i<k$ along the \emph{same trajectory}, and the actions applied between
times $i$ and $k$. This is basically the same as the 1-1 mapping and
also Amir and Russell's permutation variables. The problem with 
this definition is that the notion of state trajectory
$s,a,o,s',a',o',s'',\ldots$ is not given.}
}

\begin{definition}
A problem $P$ is \emph{causally decomposable}  when for every pair of beams $B_C(X)$ 
and $B_C(X')$ in the causal decomposition, 
then 1)~the  variables in the intersection of the two beams
are all memory variables, or 2)~there is a beam in the decomposition that includes both beams.
\end{definition}

\begin{theorem}
\label{thm:cbt:completeness}
CBT is always sound and it is complete for \emph{causally decomposable} problems.
\end{theorem}




\section{Approximation: Beam Tracking}

The last algorithm, \emph{beam tracking} is also defined over
the new \emph{causal decomposition} but it is not aimed at being
complete but rather efficient and effective. It combines the low complexity of Decoupled
CBT, which is time and space exponential in the causal width of the problem, with a
form of \emph{local consistency} that approximates the \emph{global consistency} 
enforced in   CBT:

\begin{definition}
\emph{Beam tracking}  is the belief tracking algorithm that operates on the
causal decomposition $C=\pair{T_C,B_C}$, setting the beliefs
$b_X^0$ at time $0$ over each beam $B_C(X)$, $X \in T_C$, to
the projection of the initial belief over the beam, 
and setting the successive beliefs $b_X^{i+1}$ in two steps.
First, they are set to $b_a^o$ for $b=b_X^i$, $a=a_i$, and $o=o_i$, where $a_i$ and $o_i$ are the 
action and observation at time $i$ in the execution.
Then a \emph{local} form of consistency is enforced upon these
beliefs by means of the following updates  until a fixed point is reached: 
\begin{equation}
\label{local}
b_{X}^{i+1} \ :=\ \Pi_{B_c(X)} (\, b_{X}^{i+1} \, \,  \join \, \, b_{Y}^{i+1}) \, . 
\end{equation}
\end{definition}

\noindent The filtering captured by the iterative updates in Eq.\,\ref{local}
defines a form of \emph{relational arc consistency} \cite{dechter:rac}
where equality constraints among beams sharing common variables
is  enforced  in  polynomial time and space in the size of the beams. 
Beam tracking  is \emph{sound} but \emph{not complete}.  In problems that
satisfy the conditions in Theorem~\ref{thm:cbt:completeness}, however,
the incompleteness is the sole result of replacing global consistency by local consistency. 

\Omit{The inference power of relational arc consistency
is strong, and goes well beyond most practical forms of constraint
propagation (** ref ***). \partialc runs in time and space
that are exponential in the \emph{causal width} of the problem.
As illustrated in the examples below, causal width is often much
smaller than problem width.
}

\section{Extensions}

Before considering the experiments, we  discuss briefly two simple but
useful extensions of the language and the algorithms.
The first involves  \emph{defined variables}. A variable $Z$ with domain
$D_Z$ can be defined as a function of a subset of state variables
in the problem, or as a function of the \emph{belief} over such
variables (e.g., $Z$ true when $X=Y$ is known to be true).
Variables defined in this way can then be used in action preconditions
or goals, and the immediate causes of such variables are the state
variables appearing in its definition.  All the results above carry
to domains with defined variables once they are added to the set of
target variables in the factored and causal decompositions.
The second extension is \emph{state constraints}. In Battleship, for
example, they are used to express the conditions under which the
neighbor of a cell containing a ship, will contain the same ship.
A state constraint represented by a formula $C$ can be encoded by
means of a dummy observable variable $Y$ that is always observed to
be true, and can be observed to be true only in states where $C$ is
true; i.e., $W_a(Y=true)=C$.  For the \emph{implementation}, however,
it pays off to treat such constraints $C$ as relations, and to include
them in all the `joins' over beliefs that include the variables in $C$. 
In causal belief tracking, this has no effect on the completeness or complexity of the
algorithm, but in beam tracking, changing the update in \eqref{local} to
\begin{equation}
b_{X}^{i+1} \ :=\ \Pi_{B_c(X)} (b_{X}^{i+1} \, \join \, b_{Y}^{i+1} \, \join \, C)
\label{local2}
\end{equation}
where $C$ stands for the  state constraints whose variables are included
in $B_C(X) \cup B_C(Y)$,  makes local consistency stronger with no
effect on the complexity of the algorithm.  Moreover, when there is one
such pair of beams for every state constraint, the state constraints
can increase the \emph{causal width} of the problem by a constant factor
of 2 at most, yet the \emph{effective causal width} of the problem does
not change, as the beams associated with the dummy observables introduced
for such constraints can  be ignored.

\section{Experimental Results}

We have tested the beam tracking algorithm over  three domains, 
Battleship, Minesweeper, and Wumpus,  in combination with simple 
heuristics for selecting actions. Belief tracking in these domains
is difficult \cite{minesweeper:consistency,minesweeper:inference}, and the 
sizes of the instances considered are much larger than
those used in contingent planning. Moreover, some of these domains do not have
full contingent solutions. We thus compare our on-line planner that relies on 
handcrafted heuristics with two reported solvers that rely on belief tracking
algorithms tailored to the domains. We also consider a simpler and fully-solvable 
version of Wumpus where the comparison with off-line and on-line contingent planners
is feasible. The results have been obtained on a Xeon `Woodcrest' 5140 CPU running at
2.33 GHz with 8GB of RAM.\footnote{A real-time animation for Minesweeper can be
  found at \url{https://www.youtube.com/watch?v=U98ow4n87RA}.}

\begin{table}[t]
\centering
\resizebox{.9\columnwidth}{!}{
  \begin{tabular}{rrrrrr}
    \toprule
    &&&& \multicolumn{2}{c}{avg.\ time per} \\
    \cmidrule{5-6}
               dim & policy & \#ships &       \#torpedos &      decision &          game \\
    \midrule
    $10 \times 10$ & greedy &     $4$ &     $40.0\pm6.9$ & \sci{2.4}{-4} & \sci{9.6}{-3} \\ 
    $20 \times 20$ & greedy &     $8$ &   $163.1\pm32.1$ & \sci{6.6}{-4} & \sci{1.0}{-1} \\ 
    $30 \times 30$ & greedy &    $12$ &   $389.4\pm73.4$ & \sci{1.2}{-3} & \sci{4.9}{-1} \\ 
    $40 \times 40$ & greedy &    $16$ &  $723.8\pm129.2$ & \sci{2.1}{-3} &           1.5 \\ 
    \midrule
    $10 \times 10$ & random &     $4$ &     $94.2\pm5.9$ & \sci{5.7}{-5} & \sci{5.3}{-3} \\ 
    $20 \times 20$ & random &     $8$ &   $387.1\pm13.6$ & \sci{7.4}{-5} & \sci{2.8}{-2} \\ 
    $30 \times 30$ & random &    $12$ &   $879.5\pm22.3$ & \sci{8.5}{-5} & \sci{7.4}{-2} \\ 
    $40 \times 40$ & random &    $16$ & $1,\!572.8\pm31.3$ & \sci{9.5}{-5} & \sci{1.4}{-1} \\ 
    \bottomrule
  \end{tabular}
}
\caption{Battleship. Averages over 10,000 runs. Time is in seconds.
}
\label{table:battleship:results}
\end{table}

\subsubsection{Battleship}

The problem is encoded with  6  variables per cell:\footnote{This is a
rich encoding that allows to observe when a ship has
been fully sunk.  In the experiments, this observation
was not used in order to compare with the work of \citeay{silver:uct}.}
$hit_{x,y}$ tells if a torpedo has been fired at the cell, 
$sz_{x,y}$ is  the size of the ship occupying the cell
(0 if no such ship), $hz_{x,y}$ encodes if ship is placed horizontally
or vertically, $nhits_{x,y}$ is the number of hits for ship at cell,
and $anc_{x,y}$ is the relative position of the ship on the cell.
The observable variable is  $water_{x,y}$  with  sensor model  given by 
$W_{fire(x,y)}(water_{x,y}=true)=(sz_{x,y}=0)$. In addition, state constraints 
are used to describe how a ship at a given cell constrains the variables
associated with neighboring cells. 
The goal of the problem is to achieve the equality $nhits_{x,y}=sz_{x,y}$  
over the cells that may contain a ship. In this encoding, the causal beams never contain more than
$5$ variables, even though the problem width is not bounded and grows
with the grid dimensions.

Table~\ref{table:battleship:results} shows results for
two policies: a \emph{random} policy  that fires at any non-fired cell
at random, and a \emph{greedy} policy that fires at the non-fired
cell most likely to contain a ship. Approximations of these probabilities
are obtained from the \emph{beliefs} computed  by beam tracking.
The difference in performance between the two policies
shows that the  beliefs are very informative. 
Moreover, for the $10\times10$ game, the agent fires
$40.0\pm6.9$ torpedos on average,  matching quite closely 
the results of \citeay{silver:uct} obtained with
a combination of UCT \cite{uct} for action selection, and a particle filter
\cite{russell:particle} tuned to the domain  for belief tracking.
Their approach involves 65,000 simulations per action
and  takes  in the order of 2 seconds per game, while  ours 
takes 0.0096 seconds per game.

\subsubsection{Minesweeper}

\Omit{The objective of the game is to
clear a rectangular minefield without detonating a mine by either
opening or flagging each cell in the field.
In the former case, if the cell contains a mine, the player dies,
else an integer counting the number of mines surrounding the
cell is revealed.
An initial configuration consists of a $m\times n$ field
with $k$ randomly-placed mines.}

We consider Minesweeper instances over $m \times n$ grids
containing $k$ randomly placed mines. The problem 
can be described with $3mn$ boolean state variables $mine_{x,y}$,
$opened_{x,y}$ and $flagged_{x,y}$ that denote the presence/absence
of a mine at cell $(x,y)$ and whether the cell has been opened or
flagged, and $mn$ observable variables $obs_{x,y}$ with domain $D=\{0,\ldots,9\}$.
There are two type of actions $open(x,y)$ and $flag(x,y)$ where
the first has no precondition and the second $\neg mine_{x,y}$.
The sensor model is given by formulas that specify the
integer that receives the agent when opening a cell 
in terms of the value $mine_{x',y'}$ for the surrounding cells.
The goal of the problem is to get the disjunction
$flagged_{x,y} \lor opened_{x,y}$ for each cell $(x,y)$,
provided that opening a cell with a mine
causes termination. The causal beams  contain at most 9 boolean variables,
even though the width of the problem is $3mn$.

Table~\ref{table:mines:results} shows results for
the three standard levels of the game and  a much  larger
instance. As in Battleship, the greedy policy uses
the beliefs computed by beam tracking, flagging  or opening  a 
cell when certain about its content,  else selecting the 
cell with the most extreme probability, and  flagging or opening it
according to its likely content.  
The results shown in the table are competitive with those  recently reported by 
\citeay{buffet:uct-mines}, which are achieved with a combination of UCT for action 
selection, and a domain-specific CSP solver for tracking beliefs. 
The success ratios that they report are: $80.2\pm0.48\%$ for the $8\times8$ with 10 mines,
$74.4\pm0.5\%$ for the $16\times16$ with 40 mines, and
$38.7\pm1.8\%$ for the $16\times30$ with 99 mines.
The authors do not report times, but as in Battleship,
the time required for selecting actions is likely to be orders-of-magnitude larger than the
time  taken by our algorithm.



\begin{table}[t]
\centering
\resizebox{.9\columnwidth}{!}{
  \begin{tabular}{rrrrrrr}
    \toprule
    &&&&& \multicolumn{2}{c}{avg.\ time per} \\
    \cmidrule{6-7}
               dim &     \#mines &      density & \%win &   \#guess  &      decision &      game \\
    \midrule
       $8\times 8$ &      10 &       15.6\% &  83.4 &       606  & \sci{8.3}{-3} &      0.21 \\
    $16 \times 16$ &      40 &       15.6\% &  79.8 &       670  & \sci{1.2}{-2} &      1.42 \\
    $16 \times 30$ &      99 &       20.6\% &  35.9 &     2,476  & \sci{1.1}{-2} &      2.86 \\
    $32 \times 64$ &     320 &       15.6\% &  80.3 &       672  & \sci{1.3}{-2} &      2.89 \\
    \bottomrule
  \end{tabular}
}
\caption{Minesweeper. Averages over 1,000 runs. Time is in seconds.
}
\label{table:mines:results}
\end{table}


\subsubsection{Wumpus}

An $m\times n$ instance of Wumpus \cite{russell:book} is described with known state variables
for the agent's position  and orientation, and hidden boolean
variables for each cell that tell whether there is a pit, a wumpus,
or nothing at the cell.
One more hidden state variable stores the 
position of the gold. The  observable variables are boolean:
$glitter$, $breeze_{x,y}$, $stench_{x,y}$ and $dead_{x,y}$,
with $(x,y)$ ranging over  the different cells. The actions
are  move forward, rotate right or left, and grab the gold.
The causal width for the encoding   is 6 while the problem 
width grows with the grid size. 
Table~\ref{table:wumpus:results} shows results for different
grid sizes and number of pits and wumpus for a greedy policy based on  a 
heuristic that returns the minimum cost of a safe path to the nearest
cell that may contain the gold. The beliefs computed by beam tracking
are used to determine which cells are safe (known to contain no wumpus
or pit) and may contain the gold. As the table shows, very large instances
are solved in real time. Moreover, the unsolved instances were all proven to be 
unsolvable  using  the information gathered by the agent 
and a SAT solver.

\Omit{We are not aware of any solver for Wumpus for making
a comparison, yet the figures clearly show that the beliefs
are tracked effectively and efficiently.
For instance, the $30 \times 30$ instances with 32 pits
and 32 wumpus  are solved successfully 89\% of the time,
in less than 4.4 seconds on average.
Moreover, all the unsolved instances were 
actually unsolvable, something that we proved,
using a SAT solver, from the information gathered by
the agent along the execution.
}

We also considered a simpler version of Wumpus that has been used as a benchmark 
for testing  off-line and on-line contingent planners. In this version, there are no pits, 
the agent is known to start at the bottom-left  corner, the gold is known to be at the opposite corner, 
and a wumpus is known to be on the cell below or to the left of each  diagonal cell, except for the first two 
 cells.  The problem has a full contingent solution for grids $n \times n$ for $n \ge 3$. Off-line planners  
have been shown to scale up to $n=7$ \cite{albore:ijcai09,trancao:cnf}, while on-line planners  to $n=20$ 
\cite{albore:ijcai09,shani:multi}. We  tried our K-replanner \cite{bonet:replanning} 
that is domain-independent, relies on a very effective form of belief representation 
based on literals and invariants, but is however less powerful than beam tracking.
The K-replanner manages to scale up to $n=40$. 
Beam tracking with the  heuristic above yields solutions to much larger grids,
e.g., $n > 100$, in real-time. 
Unfortunately, since it's not possible to try the other planners with the same action
selection mechanism, it is not possible to say how much of the gap in  performance is
due to the belief representation, and how much to action selection.
It must also be kept in mind that none of the planners handle action non-determinism
which poses no problem for beam tracking.

\begin{table}[t]
\centering
\resizebox{.95\columnwidth}{!}{
  \begin{tabular}{rrrrrrr}
    \toprule
    &&&&& \multicolumn{2}{c}{avg.\ time per} \\
    \cmidrule{6-7}
               dim &       \#p/\#w & \%den & \#decisions & \%win &      decision &          game \\
    \midrule
      $5 \times 5$ &     1\,/\,1 &   8.0 &      22,863 &  93.6 & \sci{3.8}{-4} & \sci{8.7}{-3} \\
    $10 \times 10$ &     2\,/\,2 &   4.0 &      75,507 &  98.3 & \sci{9.6}{-4} & \sci{7.2}{-2} \\
    $15 \times 15$ &     4\,/\,4 &   3.5 &     165,263 &  97.9 & \sci{1.6}{-3} & \sci{2.6}{-1} \\
    $20 \times 20$ &     8\,/\,8 &   4.0 &     295,305 &  97.8 & \sci{2.4}{-3} & \sci{7.2}{-1} \\
    $25 \times 25$ &   16\,/\,16 &   5.1 &     559,595 &  94.0 & \sci{3.8}{-3} &           2.1 \\
    $30 \times 30$ &   32\,/\,32 &   7.1 &     937,674 &  89.0 & \sci{4.7}{-3} &           4.4 \\
    $40 \times 40$ & 128\,/\,128 &  16.0 &   4,471,168 &   7.3 & \sci{2.8}{-3} &          12.7 \\
    $50 \times 50$ & 512\,/\,512 &  40.9 &   7,492,503 &   0.1 & \sci{1.3}{-2} &         100.4 \\
    \bottomrule
  \end{tabular}
}
\caption{Wumpus. Averages over 1,000 runs.
  Column  `\#p/\#w' refers to number of hidden pits and wumpus,
  and `\%den' to their density in the grid. Time is in seconds.
}
\label{table:wumpus:results}
\end{table}

\section{Summary}

We have  introduced  a causal decomposition scheme 
in belief tracking for planning along with a new algorithm, 
causal belief tracking, that is \emph{provably complete} for a  broad range of domains.
CBT is time exponential in the problem \emph{width},
and space exponential in the often much smaller \emph{causal width}.
%
%
We then introduced a second algorithm, beam tracking, 
as a meaningful approximation of CBT  that is 
both time and space exponential in the problem 
causal width. The empirical results in   Battleship, Minesweeper, and Wumpus,
where very  large instances are solved in real-time with a greedy policy
that is defined on top of these beliefs, suggest that the algorithm
may be practical enough,   managing to track beliefs  effectively and efficiently
over large deterministic and non-deterministic spaces. 

\subsubsection*{Acknowledgements}
This work was partially supported by EU FP7  Grant\# 270019  (Spacebook)
and MICINN CSD2010-00034 (Simulpast).
We thank Gabriel Detoni for his Tewnta framework used to implement
graphical interfaces for the three games and James Biagioni for his
command-line \texttt{wumpuslite} simulator.

\Omit{
In the future, we would to extend these ideas  to POMDPs
where belief states are not sets of states but probability distributions, and where 
not all the possible transitions and observations are equally likely.
Some general results for the probabilistic case are known \cite{boyen:dbn},
yet these do not consider the full structure of the planning problem.
}

\bibliographystyle{named-abbrv}
\bibliography{paper}

\end{document}